\documentclass[letterpaper, 10 pt, conference]{ieeeconf}  

\IEEEoverridecommandlockouts                              

\title{\LARGE \bf
VIRUS-NeRF - Vision, InfraRed and UltraSonic \\based Neural Radiance Fields
}

\author{Nicolaj Schmid$^{1,\ast}$, Cornelius von Einem$^{1,\ast}$, Cesar Cadena$^{1}$, Roland Siegwart$^{1}$, \\ Lorenz Hruby$^{2,\ast}$ and Florian Tschopp$^{3,\ast}$
\thanks{$^\ast$Authors contributed equally to this work}
\thanks{$^1$Authors are members of the Autonomous Systems Lab, ETH Zurich, Switzerland; {\tt\small \{firstname.lastname\}@mavt.ethz.ch}}
\thanks{$^2$Author is with Filics GmbH, Munich, Germany; {\tt\small hruby@filics.eu}}
\thanks{$^3$Author is with Voliro AG, Zurich, Switzerland; {\tt\small fts@voliro.com}}
\thanks{This work was supported by the ETH Mobility Initiative under the project \textit{LROD-ADAS}.}%
}

\usepackage{hyperref}
\usepackage{siunitx}
\usepackage{graphicx}
\usepackage{caption}
\usepackage{subcaption}
\usepackage{amsmath}
\usepackage[table]{xcolor}
\usepackage[nolist]{acronym}
\usepackage{cite}
\usepackage{float}
\usepackage{dblfloatfix}
\usepackage{microtype}

\definecolor{c2}{rgb}{0.761, 0.82, 0.933}
\definecolor{c1}{rgb}{0.522, 0.639, 0.871}
\definecolor{c0}{rgb}{0.282, 0.463, 0.804}

\usepackage{array}
\usepackage{multirow}
\newcolumntype{P}[1]{>{\centering\arraybackslash}p{#1}}

\begin{document}

\maketitle
\thispagestyle{empty}
\pagestyle{empty}

\begin{abstract}

Autonomous mobile robots are an increasingly integral part of modern factory and warehouse operations. 
Obstacle detection, avoidance and path planning are critical safety-relevant tasks, which are often solved using expensive LiDAR sensors and depth cameras. 
We propose to use cost-effective low-resolution ranging sensors, such as ultrasonic and infrared time-of-flight sensors by developing \textit{VIRUS-NeRF} - \textit{Vision, InfraRed, and UltraSonic based Neural Radiance Fields}.

Building upon \ac{Instant-NGP}, \textit{VIRUS-NeRF} incorporates depth measurements from ultrasonic and infrared sensors and utilizes them to update the occupancy grid used for ray marching.
Experimental evaluation in 2D demonstrates that \textit{VIRUS-NeRF} achieves comparable mapping performance to LiDAR point clouds regarding coverage.
Notably, in small environments, its accuracy aligns with that of LiDAR measurements, while in larger ones, it is bounded by the utilized ultrasonic sensors.
An in-depth ablation study reveals that adding ultrasonic and infrared sensors is highly effective when dealing with sparse data and low view variation.
Further, the proposed occupancy grid of \textit{VIRUS-NeRF} improves the mapping capabilities and increases the training speed by 46\% compared to \ac{Instant-NGP}.
Overall, \textit{VIRUS-NeRF} presents a promising approach for cost-effective local mapping in mobile robotics, with potential applications in safety and navigation tasks.
The code can be found at \href{https://github.com/ethz-asl/virus_nerf}{https://github.com/ethz-asl/virus\_nerf}.

\end{abstract}

\acresetall

\section{INTRODUCTION}

\begin{figure}[tb]
    \centering
    \includegraphics[trim={3.0cm 1.0cm 0.0cm 4.0cm}, clip, width=1.0\columnwidth]{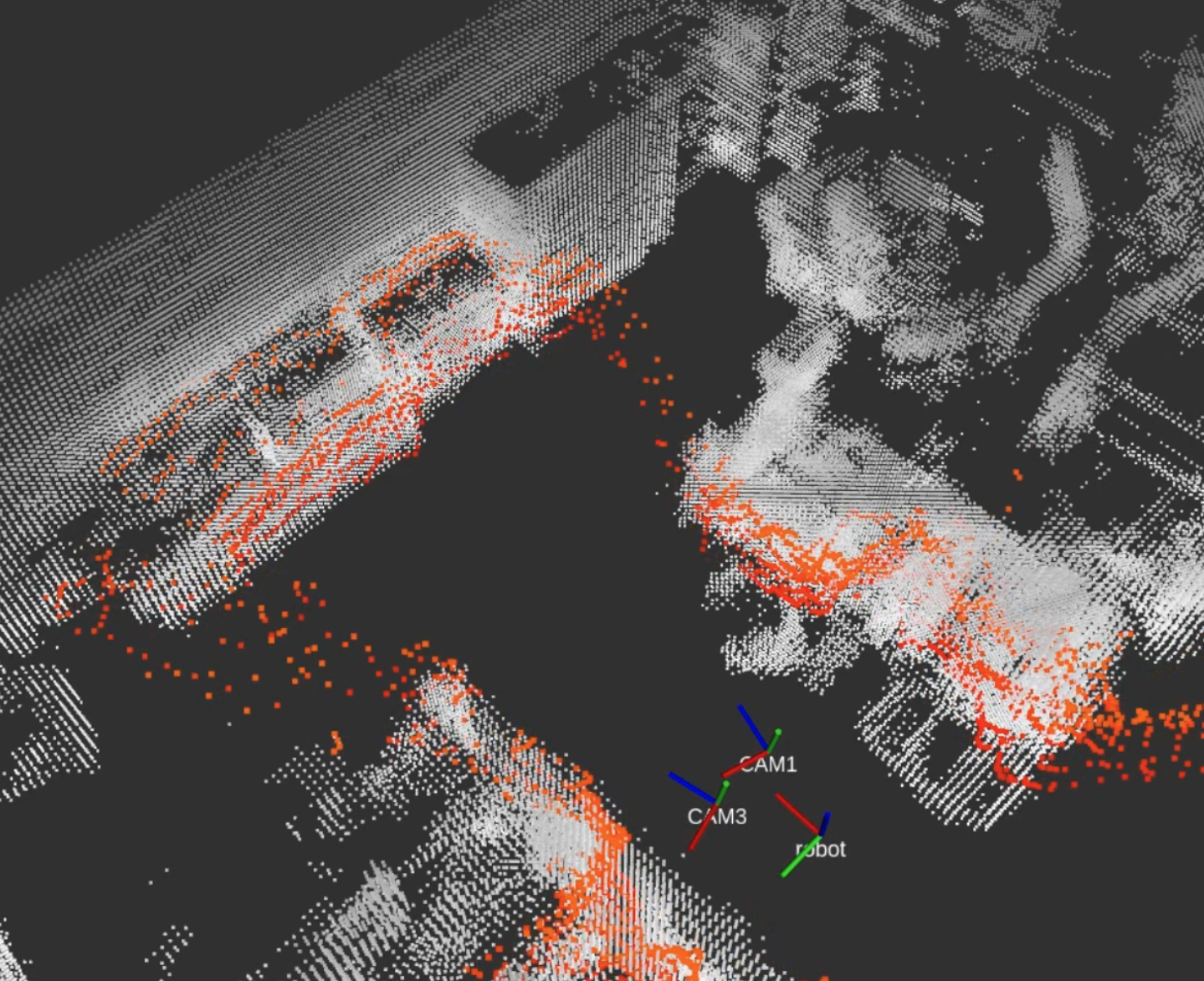}
    \caption{\textit{Office}: global map in white and \textit{VIRUS-NeRF} predictions in orange. The axes are the reference frames of the LiDAR and of the two cameras. For this visualization in 3D, \textit{VIRUS-NeRF} is inferred at multiple heights.}
    \label{fig:office_simulation}
\end{figure}

As automation advances, the demand for mobile robots is on the rise.
In the realm of factories and warehouses, \acp{AMR} exhibit remarkable flexibility and perform tasks with greater intelligence compared to traditional \acp{AGV}.
Typically, \acp{AMR} operate in semi-dynamic environments alongside human workers. 
The robot must effectively perceive its surroundings to facilitate smooth navigation and obstacle avoidance. 
Operating within the same workspace as humans requires a safe mapping algorithm, focusing particularly on the proximity of the robot (e.g. see Fig. \ref{fig:office_simulation}).
In this work, perception is studied in the simplified case of static environments and the mapping is evaluated in 2D space.

In industry, safety-critical tasks for \acp{AMR}, such as collision avoidance, often rely on costly sensors, such as 3D \ac{LiDAR} sensors.
Typically, local and instantaneous measurements are utilized being more robust and allowing faster scanning cycles than when using global mapping.
This study seeks to employ cost-efficient sensors while maintaining local mapping capabilities similar to more expensive setups.
For example, \acp{USS} are widely adapted low-cost ranging sensors in the car industry \cite{exp_disc_USS_car_parking} and many of the early mapping algorithms in mobile robotics utilize them \cite{map_og_firstPaper, map_og_bayesianUpdate, map_og_usingOGforMRperceptionAndNavigation}.
However, their low angular resolution combined with traditional fusion methods lead to sub-optimal outcomes.
Similarly, other cheap time-of-flight sensors, e.g. \acp{IRS}, result in limited mapping performances due to their sparse measurements and reduced range.
Contemporary approaches frequently rely on more advanced sensors like \acp{LiDAR} or depth cameras.
Despite their impressive performance, these setups incur substantial costs, making them less accessible for widespread adoption.
Recent advancements in stereo vision \cite{uss_stereoVision1, uss_stereoVision2} and \ac{MDE} \cite{cam_mde_review1, cam_mde_review2, cam_mde_review3} show significant progress in camera-based mapping solutions.
However, the lack of robustness, high computational requirements and the intrinsic scale ambiguity of monocular cameras remain open research challenges.
Sensor fusion may address some of these drawbacks, e.g. depth completion \cite{cam_dc_review1, cam_dc_review2, cam_dc_review3}.
Nevertheless, most of these techniques return to using costly \ac{LiDAR} or RGB-D sensors.

This study suggests using the framework of \acp{NeRF} to fuse color images with range measurements and to learn an implicit scene representation.
In the context of mobile robotics, \acp{NeRF} incur two major drawbacks: First, they converge slowly, which makes them problematic for real-time usage, and second, they require dense data with a high view variation of the environment \cite{cam_nerf_review1, cam_nerf_review3}.
More recent works address the convergence speed by proposing various improvements \cite{cam_nerf_review1, cam_dc_review2}.
Most mobile robots, especially in warehouse and factory environments, are constrained to move along pre-defined trajectories on a 2D plane, severely limiting their viewpoint variability.
Therefore, we developed \textit{VIRUS-NeRF} - a \textit{Vision, InfraRed and UltraSonic based Neural Radiance Fields}.
\textit{VIRUS-NeRF} is based on \ac{Instant-NGP}.
Similar to other works using \acp{LiDAR} \cite{cam_nerf_urban_radiance_fields, cam_nerf_cloner} or depth cameras \cite{cam_nerf_imap, cam_nerf_niceslam}, \textit{VIRUS-NeRF} complements the image-based training by depth measurements.
Notably, \textit{VIRUS-NeRF} is the first \ac{NeRF} algorithm utilizing low-cost depth sensors, i.e. \acp{USS} and \acp{IRS}.
The contributions of this work are the following:
\begin{itemize}
    \item A novel real-time, \ac{NeRF}-based sensor fusion method for integrating low-cost and low-resolution \acp{USS} and \acp{IRS} with RGB cameras. The low-cost sensors provide depth supervision, to the normally purely image-based training of \acp{NeRF}, resulting in more accurate and robust reconstructions of the environment.
    \item Improvements to the occupancy grid of \ac{Instant-NGP} using a probabilistic Bayesian formulation, which permits direct occupancy updates by depth measurements.
    \item An evaluation of \textit{VIRUS-NeRF's} mapping accuracy and coverage on real-world datasets and a direct comparison to instantaneous scans of \acp{LiDAR}, \acp{USS} and \acp{IRS}.
    \item An in-depth ablation study comparing \textit{VIRUS-NeRF} to \ac{Instant-NGP}, analyzing the contribution of the depth supervision and the improved occupancy grid separately, and studying different isolated sensor modalities.
\end{itemize}

\section{Related Work}\label{sec:related_work}
\subsection{Occupancy Grids}

One of the first mapping techniques is the occupancy grid, which is also utilized in \ac{Instant-NGP}, the base model of \textit{VIRUS-NeRF} (see section \ref{sec:related_work_nerf}).
Occupancy grids are probabilistic maps describing the occupancy state of a discretized environment~\cite{map_og_firstPaper}.
Many of the early publications explicitly use low-cost sensors, e.g. \acp{USS}~\cite{map_og_firstPaper, map_og_tesselatedProb, map_og_improvedGrid, map_og_forwardSensorModel}.
The most established implementation is based on the \textit{Bayesian Updating Rule} to integrate new measurements into the map~\cite{map_og_bayesianUpdate}.
\textit{VIRUS-NeRF} makes use of the Bayesian occupancy grid (see section~\ref{sec:occgrid}) which considers consecutive measurements and neighboring cells independently to reduce the updating complexity.
These strong assumptions may be dropped by addressing the correlation between successive samples~\cite{map_og_improvedGrid} or between neighboring cells ~\cite{map_og_forwardSensorModel}.
Multiple works extend occupancy grids to handle dynamic objects using \textit{Bayesian Occupancy Filters}~\cite{map_og_bayesian_occupancy_filter}, \textit{Particle Filters} ~\cite{map_og_particle_filter} or \textit{Markov Chains}~\cite{map_og_markov1, map_og_markov2}.

\subsection{Depth Completion}

Monocular cameras provide extremely rich information and are available for moderate prices, therefore being extensively used in mobile robotics.
\textit{VIRUS-NeRF} employs monocular cameras in the framework of \acp{NeRF}.
However, there exist some viable methods based on deep learning:
For example, \ac{MDE} predicts the pixel-wise depth by using color images~\cite{cam_mde_review2}.
RGB images have no inherent depth information and scale ambiguity limits the accuracy of \ac{MDE}.
Additionally, most \ac{MDE} algorithms do not estimate the uncertainty of their predictions~\cite{cam_mde_uncertainty1}, which makes sensor fusion difficult.
Contrary, depth completion tries to leverage the early fusion of RGB images with depth measurements.
Most approaches are based on deep learning and complement the camera with a \ac{LiDAR} sensor~\cite{cam_dc_review2}.
%
%
As an alternative to \ac{LiDAR} point clouds, radar scans can be used for depth completion~\cite{cam_dc_rad1, cam_dc_rad2, cam_dc_rad3}.
These models are relatively new because of the availability of large-scale open-source datasets containing radar data~\cite{cam_dc_radarDataSet} and the improved resolution of radar sensors in the past few years~\cite{cam_dc_rad2}.
Existing methods developed for \ac{LiDAR} do not transfer well to radars, due to the sparsity and high noise levels of the data, as well as the smaller vertical \ac{FoV}~\cite{cam_dc_rad1, cam_dc_rad3}.
To the best of our knowledge, there is no published work on utilizing \acp{USS} or \acp{IRS} for depth completion and there does not exist any large-scale open-source dataset containing such low-cost sensors.
%

\subsection{Neural Radiance Fields}\label{sec:related_work_nerf}

In 2020, Mildenhall \textit{et al.} introduce \acp{NeRF}~\cite{cam_nerf_originalPaper}.
\acp{NeRF} use \ac{MLP} to learn the geometry and the lighting of one particular three-dimensional static scene.
During inference, the model can render a new view from any position and viewing direction.
The required data is composed of images and their respective camera poses.

The original \ac{NeRF} implementation~\cite{cam_nerf_originalPaper} has some important limitations: The training of one scene takes a long time (up to $\sim$12h on a GPU~\cite{cam_nerf_review1}).
In addition, \acp{NeRF} are designed for small scenes, i.e. single objects or small rooms, and struggle with unbounded environments~\cite{cam_nerf_cloner}.
Dense data having a high view variation is required for training~\cite{cam_nerf_review1, cam_nerf_review3} and surfaces can be rugged due to a lack of geometrical constraints~\cite{cam_nerf_neus}.
Many of these drawbacks are addressed in subsequent publications, as presented below.

\ac{Instant-NGP} reduces the training time from several hours to a few minutes while achieving a similar accuracy~\cite{cam_nerf_instant_ngp}.
Instead of a sinusoidal encoding like in the original implementation, \ac{Instant-NGP} uses a multi-resolution hash encoding.
In addition, \ac{Instant-NGP} proposes to use a $128^3$ occupancy grid, making the ray marching more efficient.
The grid is updated by a heuristic rule using density predictions and thresholded by a fixed value to distinguish occupied space, where points are sampled, from unoccupied areas, which are skipped.
The current state-of-the-art in terms of surface reconstruction (including large outdoor scenes) is Neuralangelo~\cite{cam_nerf_neuralangelo}.
Neuralangelo is based on \ac{Instant-NGP} optimizing the hash grid with a coarse-to-fine approach.
Similar to NeuS~\cite{cam_nerf_neus}, Neuralangelo uses signed distance functions and is improved by adding smoothing constraints.

Urban Radiance Fields~\cite{cam_nerf_urban_radiance_fields}, CLONeR~\cite{cam_nerf_cloner}, iMAP~\cite{cam_nerf_imap} and NICE-SLAM~\cite{cam_nerf_niceslam} reduce the demand for dense data by adding depth supervision.
Urban Radiance Fields and CLONeR use \ac{LiDAR} point clouds in addition to color images.
CLONeR separates the occupancy and the color into two \acp{MLP} where the occupancy \ac{MLP} is trained with \ac{LiDAR} point clouds and the color \ac{MLP} with images.
iMAP~\cite{cam_nerf_imap} and NICE-SLAM~\cite{cam_nerf_niceslam} are \ac{SLAM} algorithms and employ depth supervision from an RGB-D camera.
More recent implementations, e.g. NeRF-SLAM~\cite{cam_nerf_nerfslam} and Orbeez-SLAM~\cite{cam_nerf_orbeezslam}, perform \ac{SLAM} utilizing only monocular cameras by separating pose estimation from neural scene representation and leveraging visual odometry.
For example, NeRF-SLAM uses Droid-SLAM~\cite{cam_nerf_droidslam} as a tracking module and \ac{Instant-NGP} for scene representation. 

\ac{NeRF} is the preferred approach in this research because of its implicit sensor fusion of color images and range measurements (see section~\ref{sec:depth_supervision}).
Meanwhile, it is a mapping framework having the following advantages: \acp{NeRF} are continuous and not discrete which allows in general a higher resolution.
Implicit scene representations are more memory-efficient than explicit ones.
For example, the smallest room with a volume of about \SI{130}{\cubic\metre} is represented by less than \SI{32}{\mega\byte} in our experiments.
Comparably, a 3D occupancy grid with \SI{1}{\cubic\centi\metre} resolution is more than 16 times larger.
Besides occupancy, \acp{NeRF} learn color and lighting properties which could be used for further tasks, e.g. object classification or scene segmentation.

\section{VIRUS-NeRF}\label{sec:virus_nerf}

\textit{VIRUS-NeRF} is based on \ac{Instant-NGP} considering its fast convergence speed and its wide adaption~\cite{cam_nerf_neuralangelo, cam_nerf_nerfslam}.
We propose two improvements on top of the base model: First, similarly to other works \cite{cam_nerf_urban_radiance_fields, cam_nerf_cloner, cam_nerf_imap, cam_nerf_niceslam}, depth supervision is added to the color-based training, reducing the demand of dense data with a high view variation.
However, instead of using expensive \acp{LiDAR} or depth cameras, \textit{VIRUS-NeRF} is based on low-cost \acp{USS} and  \acp{IRS} (see chapter \ref{sec:depth_supervision}).
Second, the occupancy grid of \ac{Instant-NGP} is updated by the depth measurements (see chapter \ref{sec:occgrid}), making ray marching more efficient and improving the results.

\subsection{Depth Supervision} \label{sec:depth_supervision}

\subsubsection{Color Rendering}
In general, \acp{NeRF} are trained as follows: During ray marching, a ray is traced in the viewing direction of every pixel.
$M$ pairs of positions and directions are sampled along this ray.
These samples are the input to a \ac{MLP} that is only a few layers deep.
The network predicts the color ($\hat{\textbf{c}}_j$) and density ($\sigma_j$) of each sample.
Then, through volume rendering, the actual color of the ray ($\hat{\textbf{C}_i}$) is estimated:
\begin{equation} \label{eq:4_volume_rendering1}
    \hat{\textbf{C}_i} = \sum_{j=1}^{M} T_j (1 - e^{-\sigma_j \delta_j}) \hat{\textbf{c}}_j
\end{equation}
where $\delta_j$ is the distance between adjacent samples and $T_j$ is the light transmittance:
\begin{equation} \label{eq:4_volume_rendering2}
    T_j = exp(- \sum_{l=1}^{j-1} \sigma_l \delta_l)
\end{equation}
Finally, the squared error between all estimated ray colors ($\hat{\textbf{C}_i}$) and the corresponding pixels ($\textbf{C}_i$) is calculated for one batch of $N$ pixels.
This loss ($\mathcal{L}_c$) is used for back-propagation:
\begin{equation} \label{eq:4_nerf_loss}
    \mathcal{L}_c = \sum_{i=1}^{N} ||\hat{\textbf{C}_i} - \textbf{C}_i ||_2^2
\end{equation}

\subsubsection{Depth Rendering}
As shown in iMAP~\cite{cam_nerf_imap}, the depth $\hat{D_i}$ of a pixel $i$ can be estimated during volume rendering:
\begin{equation} \label{eq:6_depth_rendering}
    \hat{D_i} = \sum_{j=1}^{M} \omega_j d_j = \sum_{j=1}^{M} T_j (1 - e^{-\sigma_j \delta_j}) d_j
\end{equation}
where $d_j$ is the depth of sample $j$, $\delta_j = d_{j+1} - d_j$ is the distance between adjacent samples and $T_j$ is the light transmittance (see equation~\ref{eq:4_volume_rendering2}).
The depth rendering described in equation~\ref{eq:6_depth_rendering} is equivalent to the color rendering of equation~\ref{eq:4_volume_rendering1}, except that the predicted depths $d_j$ are summed up instead of the colors $\hat{\textbf{c}}_j$.

\ac{IRS} measurements are considered to be point-like, i.e. one measurement $D_i$ corresponds to one or few camera pixels in a close neighborhood.
Analogue to the colors, the depth loss is the squared error between all estimated depths $\hat{D_i}$ and the depth measurements $D_i$:
\begin{equation} \label{eq:6_depth_loss}
    \mathcal{L}_{IRS} = \sum_{i=1}^{N} ||\hat{D_i} - D_i ||_2^2
\end{equation}

\acp{USS} have a wide opening angle and the exact location of the object reflecting the sound wave is unknown.
This prohibits applying the same depth loss as for the \ac{IRS}.
However, neglecting complete absorption and specular reflections of sound waves, an error can be calculated for all predictions that are closer than the measurement:
\begin{equation} \label{eq:6_loss_uss}
    \mathcal{L}_{USS} = \sum_{i=1}^{N} ||\hat{D_i} - D_i ||_2^2\text{, for all \textit{i} where } \hat{D_i} < D_i - \epsilon_{USS}
\end{equation}
where $\epsilon_{USS}$ corresponds to the accuracy of the \ac{USS}.
The total loss is given by the following equation:
\begin{equation} \label{eq:6_tot_loss}
    \mathcal{L}_{tot} = \mathcal{L}_{c} + \mathcal{L}_{IRS} + \mathcal{L}_{USS}
\end{equation}

\subsubsection{Rendering Bias}
\label{sec:virus_rendering_bias}

Volume rendering is a weighted sum of distances $d_j$ with weights $\omega_j = T_j (1 - e^{-\sigma_j \delta_j})$ (see equation~\ref{eq:6_depth_rendering}).
Let's assume that the densities $\sigma_j$ are described by a positive symmetric function around the surface of an object (e.g. normal distribution: center = predicted surface location, std = uncertainty).
Then, the second part of the weights $(1 - e^{-\sigma_j \delta_j})$ adopts the same symmetry as the densities.
The light transmittance $T_j$ is a monotonically decreasing function (see equation~\ref{eq:4_volume_rendering2}).
Hence, the weighting $\omega_j$ is on average larger for samples before the surface of the object than afterwards and therefore, the depth $\hat{D_i}$ is underestimated systematically.
However, the \ac{NeRF} is not tied to model symmetric density functions and the bias can be absorbed into the neural network.
Moreover, a few samples of high density may determine the depth estimation completely because the transmittance $T_j$ decreases exponentially and converges fast to zero.

\subsection{Occupancy Grid}\label{sec:occgrid}

\subsubsection{Bayesian Updating Rule}\label{sec:bayesian_updating_rule}
The occupancy grid of \textit{VIRUS-NeRF} is based on a dual updating mechanism using the \ac{NeRF} predictions similar to \ac{Instant-NGP} and additionally the depth measurements.
The key difference to \ac{Instant-NGP} is that the occupancy grid of \textit{VIRUS-NeRF} contains values in $[0,1]$ instead of $[0,\infty)$.
This allows for a probabilistic formulation and the use of the \textit{Bayesian Updating Rule}~\cite{map_og_bayesianUpdate}, which updates the probability of a cell being occupied $c_i^{occ}$ or empty $c_i^{emp}$ based on the probability $P(M_n | c_i^{occ})$ of making a measurement $M_n$:
\begin{equation}\label{eq:map_og_bayesianUpdate}
\begin{split}
    P&(c_i^{occ} |M_1, ..., M_n)  = \frac{P(M_1, ..., M_n | c_i^{occ}) P(c_i^{occ})}{P(M_1, ..., M_n)} \\
    & = \frac{P(M_n | c_i^{occ}) P(c_i^{occ} | M_1, ..., M_{n-1})}{P(M_n|c_i^{occ}) P(c_i^{occ}) + P(M_n|c_i^{emp}) P(c_i^{emp})} \\
\end{split}
\end{equation}
where $P(c_i^{occ} | M_1, ..., M_n)$ is the posterior probability given measurements $M_1, ..., M_n$ and $P(c_i^{occ} | M_1, ..., M_{n-1})$ is the occupancy grid before integrating $M_n$.
$P(c_i^{occ/emp})$ can be assumed to be equal to 0.5, or it can be derived from initial information about the map.

\subsubsection{Depth-Update}
\label{sec:depth_update}
The probability of a depth sensor making a measurement $P(M_n | c_i^{occ})$ is given by the \textit{Multiple Target Model} from the \textit{MURIEL} method~\cite{map_og_improvedGrid} proving to be the best real-time occupancy grid in a benchmark comparing different algorithms~\cite{map_og_empiricalEvaluation}.
Early-stage testing shows that the low angular resolution of \acp{USS} prevent the occupancy grid from refining.
Therefore, the \textit{Depth-Update} is done uniquely based on \ac{IRS} measurements.

\subsubsection{NeRF-Update}\label{sec:nerf_update}
The \textit{NeRF-Update} uses no conventional measurement.
However, the term $P(M_n=\sigma_i | c_i^{occ})$ can be thought of as the probability of the \ac{NeRF} predicting the density $\sigma_i$ given that a cell is occupied.
The \ac{MLP} outputs density predictions between zero and infinity.
To be able to use the \textit{Bayesian Updating Rule}, we introduce the projection of the density from $[0, \infty)$ to $[0, 1]$:
\begin{equation} \label{eq:6_occgrid0}
    P(M_n=\sigma_i | c_i^{occ}) = \frac{1}{1 + (\frac{\sigma_T}{\sigma_i})^\zeta}
\end{equation}
where $\sigma_i$ is the predicted density by the \ac{NeRF}, $\zeta$ is the slope of the mapping function and $\sigma_T$ is the density threshold.
If $\zeta \longrightarrow \infty$, then the projection $P(M_n=\sigma_i | c_i^{occ})$ becomes a step function.
If $\sigma_i > \sigma_T$, then the occupancy probability $P(M_n=\sigma_i | c_i^{occ})$ is larger than $0.5$ and vice versa.
The weights of the density \ac{MLP} are randomly initialized in $[-\frac{1}{\sqrt{32}}, \frac{1}{\sqrt{32}}]$ leading to small $\sigma_i$ values at the beginning of the training.
If $\sigma_T$ is fixed, then $P(M_n=\sigma_i | c_i^{occ})$ would vanish in the first few cycles. Therefore, $\sigma_T$ is defined as a function of $\sigma_i$ as follows: $\sigma_T = \min (\sigma_{Tmax}, \frac{1}{N} \sum_{i=1}^{N}\sigma_i)$ where $\sigma_{Tmax}$ is the maximum density threshold and $N$ is the batch size.
For all tested models, $\sigma_{T}$ becomes constant after a few training steps, s.t. $\sigma_T = \sigma_{Tmax}$.

\section{Experiments}\label{sec:experiments}

\subsection{Implementation}
In this work, the \textit{Taichi implementation}\footnote{\href{https://github.com/Linyou/taichi-ngp-renderer}{https://github.com/Linyou/taichi-ngp-renderer}} \cite{taichi} is employed because the original implementation is written in \textit{Cuda} and therefore does not run on a CPU which is required for development.

\subsection{Dataset}

\subsubsection{Environment}
The dataset is collected by a mobile robot in two environments: \textit{Office} (room with tables, chairs and cupboards, \SI{72}{\metre\squared}) and \textit{Common Area} (room with tables, couches and kitchen corner, \SI{216}{\metre\squared}).
The scenes are captured quasi-statically containing only minimal movements, e.g. a person writing on a keyboard.
The robot explores the environments on 2D trajectories and makes the measurements with sensors having an overlapping \ac{FoV}.
This leads to a relatively low view variation of the scene.

\begin{figure}
  \begin{subfigure}[b]{0.383\columnwidth}
    \includegraphics[trim={0.0cm 0.0cm 0.0cm 0.0cm}, clip, width=\linewidth]{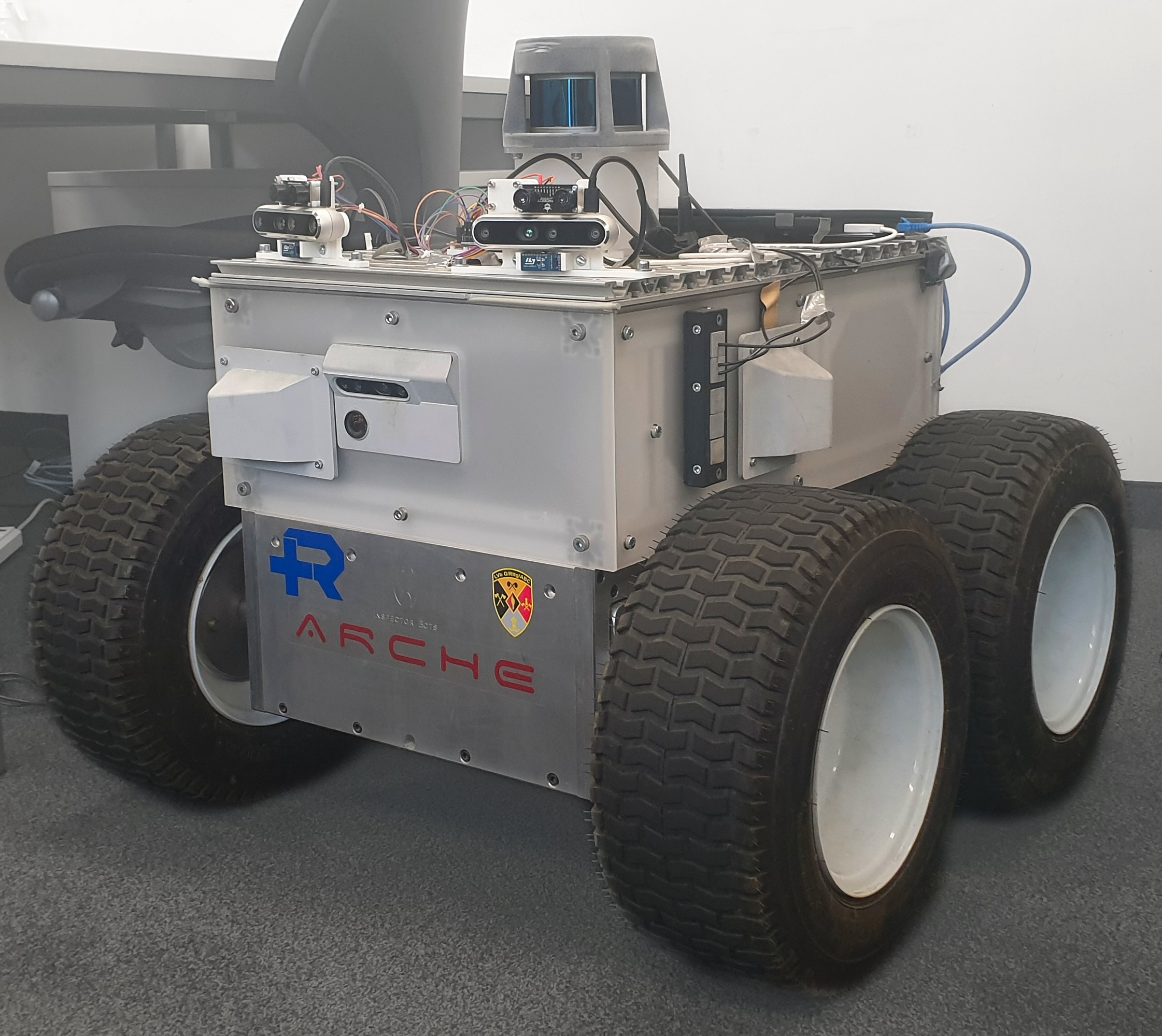}
    \caption{Super Mega Bot}
    \label{fig:robot}
  \end{subfigure}
  \hfill 
  \begin{subfigure}[b]{0.597\columnwidth}
    \includegraphics[trim={0.0cm 0.0cm 0.0cm 0.0cm}, clip, width=\linewidth]{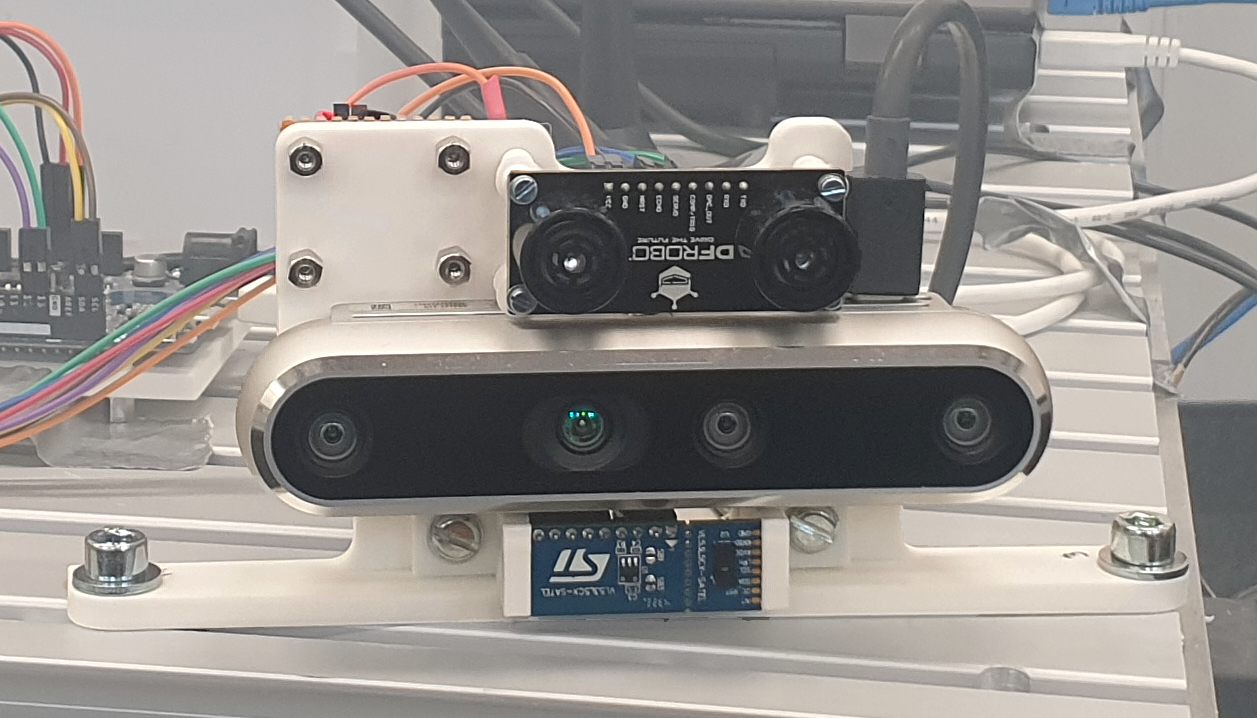}
    \caption{Sensor stack: USS, camera and IRS}
    \label{fig:sensor_stack}
  \end{subfigure}
\caption{Experimental setup}
\label{fig:setup}
\end{figure}

\subsubsection{Hardware}
The \ac{SMB} is used - a differential drive robot designed by \textit{Inspector Bots}\footnote{\href{https://www.inspectorbots.com/}{https://www.inspectorbots.com/}} (see Fig.~\ref{fig:robot}).
On top of the \ac{SMB}, a \textit{RoboSense RS-LiDAR-16} and two sensor stacks are fixed.
The sensor stack mounts the \ac{USS} \textit{DFRobot URM37}\footnote{\href{https://wiki.dfrobot.com/URM37_V5.0_Ultrasonic_Sensor_SKU_SEN0001_}{https://wiki.dfrobot.com/URM37\_V5.0\_Ultrasonic\_Sensor\_SKU\_SEN0001\_}}, the \ac{IRS} \textit{STMicroelectronics VL53L5CX}\footnote{\href{https://www.st.com/en/imaging-and-photonics-solutions/vl53l5cx.html}{https://www.st.com/en/imaging-and-photonics-solutions/vl53l5cx.html}} and the camera \textit{Intel RealSense D455}\footnote{\href{https://www.intelrealsense.com/depth-camera-d455/}{https://www.intelrealsense.com/depth-camera-d455/}} (see Fig.~\ref{fig:sensor_stack}).
They are designed such that all sensors of one stack are oriented in the same direction and are as close as possible.
The \ac{IRS} is a time-of-flight sensor measuring an array of 8x8 pixels.
The stacks are approximately \SI{27}{\centi\metre} apart and point in $\pm \SI{14.5}{\degree}$ to the driving direction.
The configuration of the cameras having an overlapping \ac{FoV} is chosen to use the calibration tool \textit{Kalibr} \cite{exp_kalibr1} which estimates the intrinsic and extrinsic camera parameters simultaneously.
The calibration concerning the \ac{LiDAR} is done using the \textit{Camera-LiDAR Calibration - V 2.0} \cite{exp_cam2lidar_calibration}.
The influence of an orientation error between the IRS and the camera is tested by simulating \ac{IRS} measurements and adding an artificial angular error.
The \ac{NND} is not affected by an error up to \SI{3}{\degree} and therefore, the \ac{IRS} and \ac{USS} calibration obtained from \ac{CAD} is sufficient.
The sensors are not hardware-synchronized but recorded on the same clock and the temporally closest measurements are assigned during post-processing for each sensor stack.

\subsubsection{Localisation}
The \ac{NeRF} requires the measurement poses for training.
Therefore, the poses are estimated by \textit{KISS-ICP} based on the \ac{LiDAR} point clouds~\cite{exp_process_kiss_icp} and optimized with~\cite{exp_opt_balm} a bundle adjustment for \ac{LiDAR} mapping.

\subsection{Metrics}

Traditionally, safety-relevant navigational tasks are done using instantaneous depth measurements.
Therefore, \textit{VIRUS-NeRF} is compared to momentary scans of \acp{USS}, \acp{IRS} and \acp{LiDAR}.
For every scene, a global map is created by projecting the \ac{LiDAR} point clouds to a grid in world coordinates using the optimized poses.
This global map has a \SI{3}{\cubic\centi\metre} cube size and is thresholded at a minimum of two points per voxel to reduce noise.
For every test point, the \ac{GT} consists of a \SI{360}{\degree} 2D depth scan at the height of the cameras within the global map.
The depth predictions are not compared directly to the global map to mitigate the erroneous association of objects. 
For example, a depth prediction can be too far away, but another object is present at this location in the global map. This would lead to a small error even though the prediction is false.
For \textit{VIRUS-NeRF}, the predictions consist of a \SI{360}{\degree} 2D depth scan at the height of the cameras, which is created by volume rendering (see equation~\ref{eq:6_depth_rendering}).
For the depth sensors, the predictions are obtained by collapsing the measurements to a 2D representation in a vertical range of $\pm \SI{5}{\centi\metre}$ above and below the camera height. 

To compare the scans, the \ac{NND} is calculated which is less sensitive to orientation errors than the \ac{RMSE}.
The \ac{NND} can be calculated in two directions: The distance from every prediction point to the closest \ac{GT} point is a measure of accuracy.
The inverse direction describes the coverage of the \ac{GT} by the prediction.
The \ac{NND} is given by the mean of all points and an inlier-outlier metric.
Inliers are defined as points where the \ac{NND} is less than \SI{10}{\centi\metre}.
The assessment of accuracy, coverage, and inlier ratio aligns with the evaluation criteria employed in iMAP~\cite{cam_nerf_imap} and NICE-SLAM~\cite{cam_nerf_niceslam}, wherein the mapping coverage is denoted as \textit{completion} and the proportion of inliers as \textit{completion ratio}, using a threshold of \SI{5}{\centi\metre} instead of \SI{10}{\centi\metre}.
%
%
%
%
Finally, all metrics are determined for three zones defined by the \ac{GT} depth.
The zones roughly represent different applications: The first zone ($0-\SI{1}{\metre}$) concerns safety applications, the second one ($0-\SI{2}{\metre}$) tasks like obstacle detection and the third one ($0-\SI{100}{\metre}$) path planning.

\subsection{Mapping}

\subsubsection{Results}
\begin{figure}[bt]
    \centering
    \includegraphics[trim={0.2cm 0.3cm 0.3cm 1.2cm}, clip, width=1.0\columnwidth]{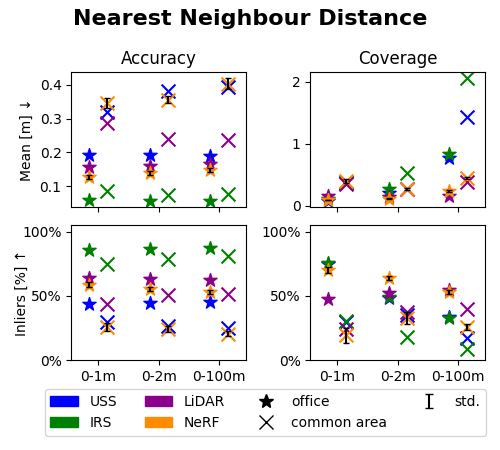}
    \caption{\textit{Office} and \textit{Common Area} NND: The first column describes the accuracy and the second one the coverage. The rows show the mean NND and the inlier ($NND < \SI{10}{\centi\metre}$) percentage. Each metric is calculated for three zones defined by the \ac{GT} depth. \textit{VIRUS-NeRF} is evaluated for 10 runs and the error bar indicates the standard deviation.}
    \label{fig:office_metrics}
\end{figure}

The average statistics for all test points and over 10 runs are summarized in Fig. \ref{fig:office_metrics} for the \textit{office} and the \textit{common area} environment.
While the amplitude of the metric depends on the particular environment, the tendency is everywhere likewise: The \ac{USS} has the worst accuracy of all sensors.
Up to zone 2 ($0-\SI{2}{\metre}$), its coverage is close to the one of the \ac{LiDAR} due to its large opening angle.
However, in the third zone ($0-\SI{100}{\metre}$), the coverage of the \ac{USS} worsens significantly.
The \ac{IRS} achieves the best accuracy while having the worst coverage due to sparse measurements.
The inferior accuracy of the \ac{LiDAR} sensor compared to the \ac{IRS} cannot be explained by the higher range of the \ac{LiDAR} sensor because it is also present in zones 1 and 2.
\acp{LiDAR} retain the best coverage in the third zone ($0-\SI{100}{\metre}$).

\begin{figure}[b]
  \begin{subfigure}[b]{0.49\columnwidth}
    \includegraphics[trim={1.5cm 0.4cm 15.8cm 19.3cm}, clip, width=\linewidth]{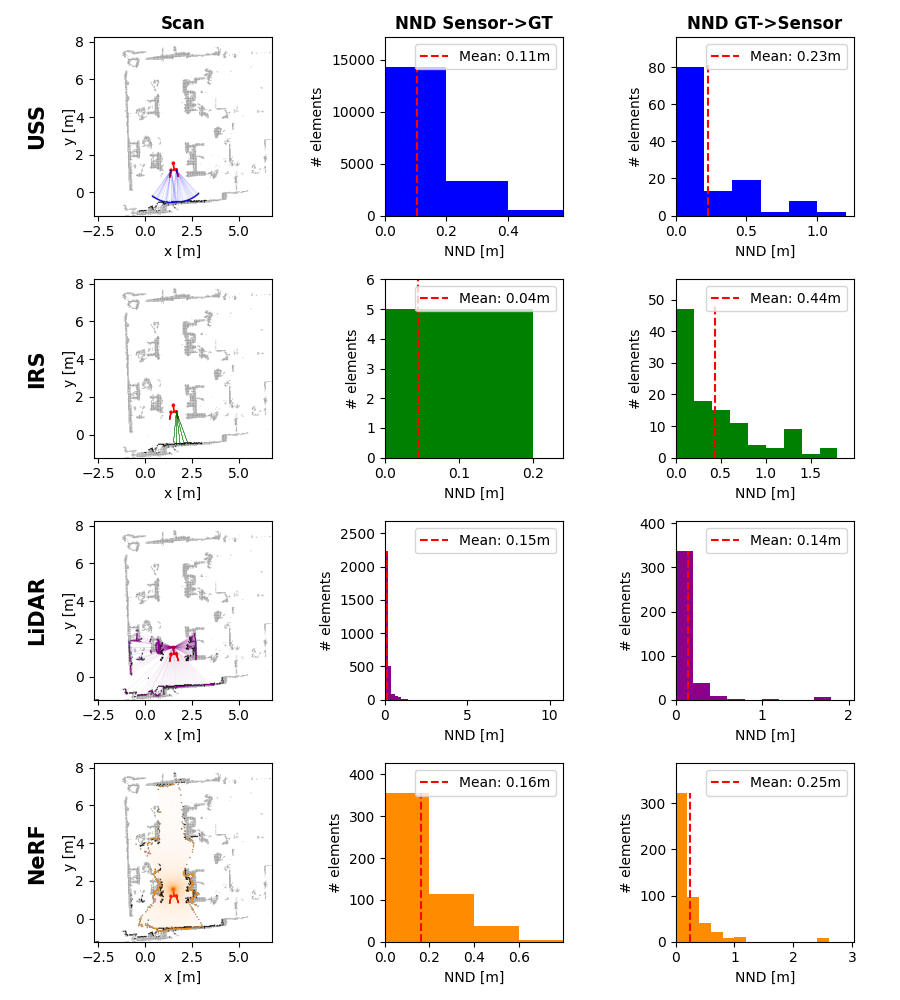}
    \caption{Test point 37}
    \label{fig:office_map37}
  \end{subfigure}
  \hfill 
  \begin{subfigure}[b]{0.49\columnwidth}
    \includegraphics[trim={1.5cm 0.4cm 15.8cm 19.3cm}, clip, width=\linewidth]{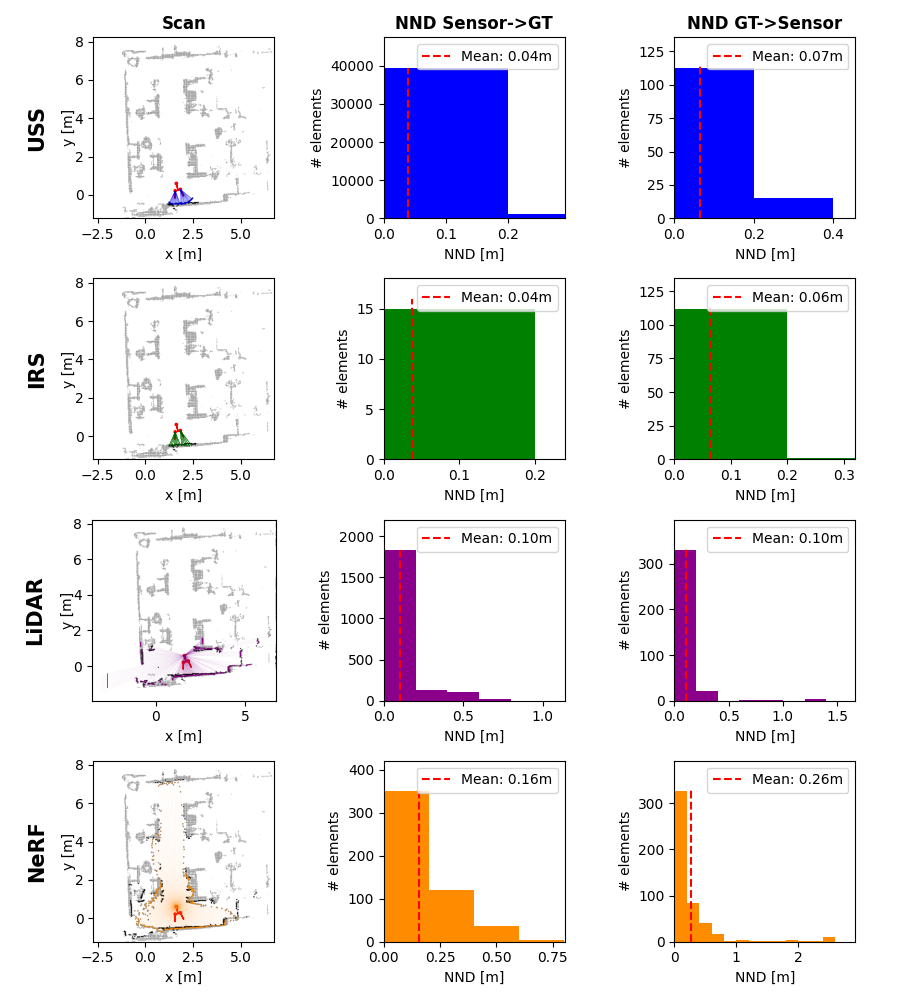}
    \caption{Test point 44}
    \label{fig:office_map44}
  \end{subfigure}
\caption{\textit{Office}: robot in red, global map in grey, \ac{GT} scan in black and \textit{VIRUS-NeRF} (USS, IRS \& camera) in orange.}
\label{fig:office_maps}
\end{figure}

\textit{VIRUS-NeRF} scores a comparable coverage than the \ac{LiDAR}.
The accuracy of the \ac{NeRF} depends on the scene: For smaller scenes, e.g. the \textit{office}, it is slightly better than \acp{LiDAR} but for larger ones, e.g. the \textit{common area}, it exhibits performance akin to \ac{USS}.
The outliers ($NND > \SI{10}{\centi\metre}$) can be separated into predictions that are \textit{too close} to the robot or \textit{too far} away relative to the \ac{GT}.
Analyzing this distinction for the coverage in zone three ($0-\SI{100}{\metre}$) shows that the largest part of all predictions is \textit{too close} with approximately three-quarters of all outliers in the \textit{common area} and 90\% in the \textit{office}.
This trend is also visible in Fig.~\ref{fig:office_maps}: \textit{VIRUS-NeRF} tends to underestimate the distance to objects, leading to false-positive predictions.

\subsubsection{Discussion}

The results reflect the theoretical advantages and weaknesses of the sensors: The accuracy of the \ac{USS} is limited by the poor angular resolution.
The acceptable coverage at short range is in line with common \ac{USS} applications where coverage is more important than accuracy, e.g. for car parking assistants \cite{exp_disc_USS_car_parking}.
The \acp{IRS} have an impressive accuracy, being at least two orders of magnitude cheaper than the \ac{LiDAR} sensor.
However, the \ac{LiDAR} sensor seems to be the best trade-off between accuracy and coverage.
%
%

The reduced accuracy of \textit{VIRUS-NeRF} in the \textit{common area} compared to the \textit{office} may be explained by the \ac{IRS} having a short range of \SI{4}{\metre}. 
In the \textit{office}, 35.7\% of all \ac{IRS} measurements are valid compared to only 10.3\% in the \textit{common area} where the objects are more spread out.
However, for safety-relevant tasks, e.g. obstacle detection, the coverage is more important than accuracy and this one remains comparable to \ac{LiDAR} point clouds in all environments.
\textit{VIRUS-NeRF} predicts more outliers \textit{too close} leading to false-positive predictions.
When visualizing the results as in Fig.~\ref{fig:office_maps}, the estimation in front of the robot is usually accurate and the hallucinations are either sideways to the trajectory (e.g. $x=\SI{0}{\metre}, y=\SI{0}{\metre}$ in Fig.~\ref{fig:office_map44}) or further away from the current position of the robot (e.g. $x=\SI{2.5}{\metre}, y=\SI{3}{\metre}$ in Fig.~\ref{fig:office_map44}).
Lateral to the path only a few measurements are taken causing erroneous predictions.
This could be addressed by adding sensors pointing in these directions.
Distant hallucinations are most likely caused by the volume rendering, which is biased towards underestimating the depth as explained in chapter~\ref{sec:virus_rendering_bias}.
If the robot moves closer to a particular region, then fewer dispensable samples influence the volume rendering and the false-positive predictions disappear, e.g. compare $(x=\SI{2.5}{\metre}, y=\SI{0}{\metre})$ in Fig.~\ref{fig:office_map37} and \ref{fig:office_map44}.
In contrast to path planning, hallucinations are not as critical for safety-relevant tasks, e.g. collision avoidance, especially if they vanish when moving closer.

\subsection{Ablation Study}
\subsubsection{Results}

\begin{figure}[b]
  \begin{subfigure}[b]{0.49\columnwidth}
    \includegraphics[trim={2.1cm 0.4cm 15.3cm 19.3cm}, clip, width=\linewidth]{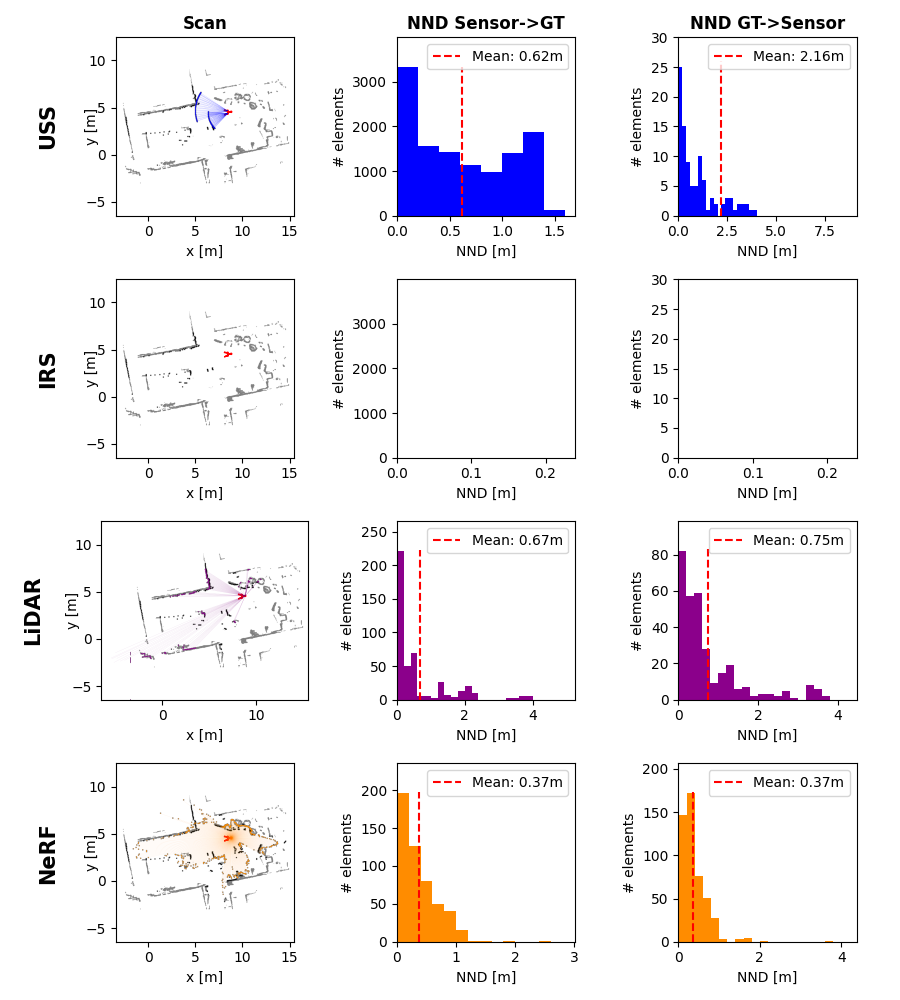}
    \caption{USS, IRS \& camera}
    \label{fig:commonroom_map23}
  \end{subfigure}
  \hfill 
  \begin{subfigure}[b]{0.49\columnwidth}
    \includegraphics[trim={2.1cm 0.4cm 15.3cm 19.3cm}, clip, width=\linewidth]{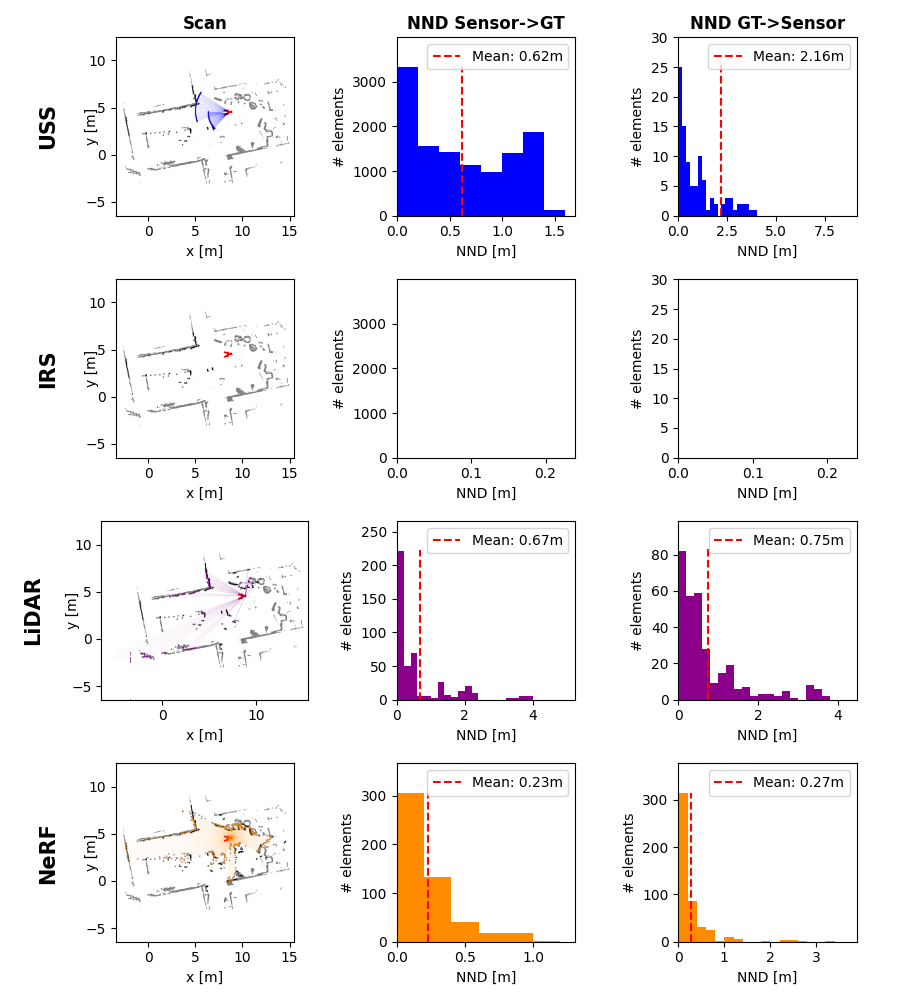}
    \caption{RGB-D camera}
    \label{fig:commonroom_map23_rgbd}
  \end{subfigure}
\caption{\textit{Common Area} test point 23: robot in red, global map in grey, \ac{GT} scan in black and \textit{VIRUS-NeRF} in orange.}
\label{fig:commonroom_maps}
\end{figure}

In the ablation study, \textit{VIRUS-NeRF} is compared to \ac{Instant-NGP}. Additionally, the contribution of the depth supervision and the improved occupancy grid are studied, and different sensor modalities are analyzed.
\textit{VIRUS-NeRF} is developed and the hyper-parameters are fine-tuned in the \textit{office} environment and the results of the \textit{common area} show if the model generalizes well.
The mean and inlier percentage are shown in table \ref{tab:ablation_commonarea}.
For the \textit{common area}, the RGB-D camera outperforms all other sensor constellations because its depth images are denser and more accurate than low-cost alternatives (compare Fig.~\ref{fig:commonroom_map23} and \ref{fig:commonroom_map23_rgbd}).
The second-best results are achieved by using \textit{VIRUS-NeRF}.
When removing either the \ac{USS} or the \ac{IRS}, the performance drops.
The original \ac{Instant-NGP} implementation is significantly worse in all metrics.
When adding the depth losses and still using the occupancy grid of \ac{Instant-NGP}, the results improve but do not reach the same level as \textit{VIRUS-NeRF}.
The same ablation study is repeated for the \textit{office} environment: The results are similar, with the exception that omitting pose optimization is less severe or even better in terms of inlier metrics than in the \textit{common area}, as smaller scenes are less affected by odometry drift.
The occupancy grid of \ac{Instant-NGP} has slightly better mean coverage than \textit{VIRUS-NeRF} in the \textit{office}.

\begin{table}[]
    \centering
    \begin{tabular}{|| P{2.0cm} P{0.6cm} || P{0.75cm} P{0.75cm}  | P{0.75cm} P{0.75cm} ||} \hline\hline
        \textbf{Occ. Grid} & & \multicolumn{2}{c|}{\textbf{Mean [m] $\downarrow$}}  & \multicolumn{2}{c||}{\textbf{Inliers [\%] $\uparrow$}} \\
        \textbf{/ Sensors} & \textbf{Scene} & Acc. & Cov. & Acc. & Cov. \\ \hline\hline
        
         Instant-NGP \cite{cam_nerf_instant_ngp} & C & 0.712 & 1.287 & 0.056 & 0.059 \\ 
         CAM & \textit{O} & \textit{0.281} & \textit{0.497} & \textit{0.221} & \textit{0.201} \\ \hline
         Instant-NGP & C & 0.509 & 0.501 & 0.131 & 0.22  \\ 
        CAM, USS, IRS & \textit{O} & \textit{0.164} & \cellcolor{c2}\textit{0.214} & \textit{0.497} & \textit{0.506} \\ \hline\hline 
        VIRUS-NeRF & C & 0.704 & 1.412 & 0.052 & 0.056  \\ 
        CAM & \textit{O}  & \textit{0.277} & \textit{0.476} & \textit{0.231} & \textit{0.215} \\ \hline
        VIRUS-NeRF & C & 0.49 & 0.568 & 0.146 & 0.146 \\
         CAM, USS & \textit{O} & \textit{0.262} & \textit{0.349} & \textit{0.212} & \textit{0.176} \\ \hline
        VIRUS-NeRF & C & 0.625 & 0.643 & 0.097 & 0.169 \\ 
         CAM, IRS & \textit{O} & \textit{0.23} & \textit{0.374} & \textit{0.415} & \textit{0.46} \\ \hline
        VIRUS-NeRF & C & \cellcolor{c1}0.324 & \cellcolor{c1}0.378 & \cellcolor{c1}0.324 & \cellcolor{c1}0.389 \\ 
         RGB-D & \textit{O} & \cellcolor{c2}\textit{0.154} & \cellcolor{c1}\textit{0.139} &  \cellcolor{c1}\textit{0.575} & \cellcolor{c1}\textit{0.633} \\ \hline \hline
         VIRUS-NeRF & C & 0.728 & 0.922 & 0.113 & 0.186 \\ 
        Poses not opt. & \textit{O} & \textit{0.168} & \textit{0.238} & \cellcolor{c2}\textit{0.531} & \cellcolor{c2}\textit{0.554} \\ \hline
        \textbf{VIRUS-NeRF} & C & \cellcolor{c2}0.403 & \cellcolor{c2}0.448 & \cellcolor{c2}0.206 & \cellcolor{c2}0.256 \\ 
        \textbf{CAM, USS, IRS} & \textit{O}  & \cellcolor{c1}\textit{0.148} & \textit{0.237} & \textit{0.528} & \textit{0.531} \\ \hline \hline 
    \end{tabular}
    \caption{Ablation Study: The accuracy and coverage of the NND is calculated for zone 3 ($0-\SI{100}{\metre}$) and averaged over 10 runs. The first column shows which occupancy grid and sensors are used for training (CAM is an RGB camera) and the second one the scene of interest (C for the \textit{Common Area} and \textit{O} for the \textit{Office}). The first row is \ac{Instant-NGP} \cite{cam_nerf_instant_ngp} and the last one \textit{VIRUS-NeRF}. Row 7 results from \textit{VIRUS-NeRF} when not optimizing the poses with \cite{exp_opt_balm}.}
    \label{tab:ablation_commonarea}
\end{table}

\subsubsection{Discussion}
It may be surprising that mapping without depth supervision produces poor results while standard \ac{NeRF}~\cite{cam_nerf_originalPaper} as well as \ac{Instant-NGP}~\cite{cam_nerf_instant_ngp} are solely based on images showing good results in their respective studies, and other variants can represent large environments~\cite{cam_nerf_neuralangelo}.
However, the dataset of this project reflects a real robot trajectory and therefore sparse measurements and a low view diversity compared to most other datasets.
For example, iMAP~\cite{cam_nerf_imap}, NICE-SLAM~\cite{cam_nerf_niceslam} and NeRF-SLAM~\cite{cam_nerf_nerfslam} are assessed on the indoor scenes of the \textit{Replica} dataset~\cite{cam_nerf_replica_dataset} sampling twice as many images along a random trajectory.
Therefore, the images of \textit{Replica} are characterized by significantly greater variations in viewing perspectives. 

\subsection{Training}

\begin{figure}
    \centering
    \begin{subfigure}[b]{0.49\columnwidth}
        \includegraphics[trim={0.3cm 0.4cm 0.4cm 0.3cm}, clip, width=\linewidth]{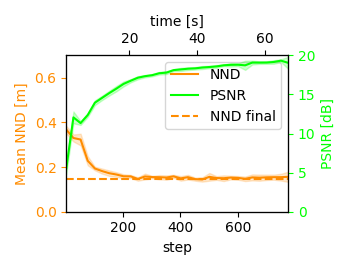}
        \caption{Offline Training}
        \label{fig:office_offline}
      \end{subfigure}
      \hfill 
      \begin{subfigure}[b]{0.49\columnwidth}
        \includegraphics[trim={0.3cm 0.4cm 0.4cm 0.3cm}, clip, width=\linewidth]{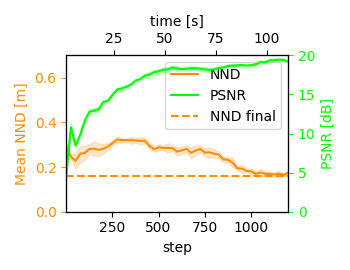}
        \caption{Online Training}
        \label{fig:office_online}
      \end{subfigure}
    \caption{\textit{Office}: The colored area indicates the standard deviation over 10 runs. The NND is the accuracy in zone 3 ($0-\SI{100}{\metre}$).}
    \label{fig:office_training}
\end{figure}

All training is done on a \textit{Nvidia Titan Xp} GPU.
The algorithm is expected to be approximately 20\% faster if using the \textit{Cuda} implementation of \ac{Instant-NGP} instead of the \textit{Taichi} implementation utilized in this study \cite{taichi}.

\subsubsection{Speed}
Fig.~\ref{fig:office_offline} shows the average \ac{NND} over 10 runs for the \textit{office} environment during training.
Using the entire dataset, \textit{VIRUS-NeRF} converges after around \SI{20}{\second}.
On average, \textit{VIRUS-NeRF} makes 11.64 training steps per second, in contrast to 7.96 when using the grid of \ac{Instant-NGP}.
This is a speed-up of 46\% and can be explained by two factors: \ac{Instant-NGP} samples during the first 256 training steps all $128^3$ grid cells and afterwards a quarter, which is significantly more than 1024 samples used in \textit{VIRUS-NeRF}.
Additionally, the occupancy grid of \textit{VIRUS-NeRF} relies partially on probabilistic sensor models (i.e. \textit{Depth-Update}, see section~\ref{sec:depth_update}) instead of inferring with the \ac{NeRF} network which is computationally less expensive.
Accelerating the training process with the occupancy grid of \textit{VIRUS-NeRF} could be very interesting for any real-time application based on depth supervised  \ac{Instant-NGP}, e.g. NeRF-SLAM~\cite{cam_nerf_nerfslam}.
To further increase the convergence speed, the same forward pass could be used for the \textit{NeRF-Update} of the occupancy grid as for the training of the \acs{MLP} instead of treating these tasks separately.
Furthermore, the density \ac{MLP} could be partially trained without the color \ac{MLP} leveraging the existence of depth measurements.

\subsubsection{Offline vs. Online}
Up to here, all results are generated in offline mode where all the data is available from the beginning of the training.
However, in most real-world applications the mapping algorithm should already be functional before all data is collected.
Online operation is using uniquely the measurements that would be available up to this time point and data playback is performed in real-time.
In this case, the training lasts for the duration of the experiment and is evaluated exclusively on already visited poses.
For online operation, the \ac{NND} converges after \SI{90}{\second} which is significantly slower than after \SI{20}{\second} during offline training (see Fig. \ref{fig:office_training}).
Similarly, the \ac{PSNR} takes online much longer to converge.
The final \ac{NND} for online learning is \SI{0.161}{\metre} compared to \SI{0.148}{\metre} in the offline case.
The \textit{common area} has similar results.
However, the metrics worsen again towards the end of the experiment.

The computational speed and the training data are possible limitations of the convergence speed.
It seems that the available data is more important compared to the computational power because, despite making more training steps, the online algorithm converges much slower than the offline one.
In the online training of the \textit{office} environment, the \ac{NND} starts to converge after half of the training process being approximately the moment when the robot turns around and drives back towards its starting position.
Similarly, the metrics degrade in the \textit{common area} when the robot makes a sharp turn and starts moving to an unseen part of the scene.
These observations and the general performance difference between offline and online training suggest that a higher variety of viewpoints would improve the convergence significantly.
This is in line with most \ac{NeRF} algorithms that rely on a plurality of viewing angles~\cite{cam_nerf_review1, cam_nerf_review3} and could be addressed by adding more sensors to the robot pointing sideways and backwards.

\section{Conclusion}\label{sec:conclusion}

This study presents \textit{VIRUS-NeRF} - \textit{Vision, InfraRed and UltraSonic based Neural Radiance Fields} for local mapping.
\textit{VIRUS-NeRF} utilizes low-cost \acp{USS} and \acp{IRS} by adapting the depth supervision to sensors having a poor angular resolution.
Additionally, the algorithm uses the sensors to update the occupancy grid introduced in \ac{Instant-NGP}~\cite{cam_nerf_instant_ngp} which is used for ray marching.
Two datasets are collected to evaluate the algorithm and to compare it to instantaneous scans of \acp{USS}, \acp{IRS} and \acp{LiDAR} in 2D.
The results show that \textit{VIRUS-NeRF} has comparable coverage to \acp{LiDAR} and is much better than \acp{USS} and \acp{IRS}.
The accuracy of \textit{VIRUS-NeRF} depends on the environment: For smaller scenes (\textit{office}), it is slightly better than \acp{LiDAR} but for larger ones, (\textit{common area}), it exhibits accuracy akin to \ac{USS}.
Larger environments could be addressed by taking a \ac{IRS} having a wider range.

The ablation study shows that the base model \ac{Instant-NGP} has substantially worse results compared to \textit{VIRUS-NeRF}.
Adding the \ac{USS} and the \ac{IRS} to the RGB image-based training is very effective, even though the low-cost sensors have a poor angular resolution and make sparse measurements respectively.
Only the more expensive RGB-D camera outperforms this sensor configuration.
The occupancy grid of \textit{VIRUS-NeRF} improves the metrics compared to the one of \ac{Instant-NGP} and it makes the algorithm 46\% faster.
Generally, the convergence speed and accuracy could be improved by adding more sensors and taking measurements with a higher view variation.
This research shows that \textit{VIRUS-NeRF} is an effective method for local mapping based on a low-cost sensor setup.

\begin{acronym}
    \acro{AGV}{Automated Guided Vehicle}
    \acro{AMR}{Atonomous Mobile Robot}
    \acro{ASL}{Autonomous Systems Lab}
    \acro{FoV}{Field of View}
    \acro{GT}{Ground Truth}
    \acro{IMU}{Inertial Measurement Unit}
    \acro{Instant-NGP}{Instant Neural Graphics Primitives with a Multiresolution Hash Encoding}
    \acro{ICP}{Iterative Closest Point}
    \acro{IRS}{Infrared Sensor}
    \acro{LiDAR}{Light Detection And Ranging}
    \acro{MDE}{Monocular Depth Estimation}
    \acro{MLP}{Multi-Layer Perceptrons}
    \acro{NeRF}{Neural Radiance Field}
    \acro{NND}{Nearest Neighbour Distance}
    \acro{PSNR}{Peak Signal to Noise Ratio}
    \acro{RMSE}{Root Mean Square Error}
    \acro{RGB-D}{Red Green Blue Depth}
    \acro{ROS}{Robot Operating System}
    \acro{SLAM}{Simultaneous Localization and Mapping}
    \acro{SMB}{Super Mega Bot}
    \acro{USS}{Ultrasonic Sensor}
    \acro{CAD}{Computer Aided Design}
\end{acronym}

\bibliographystyle{ieeetran}
\bibliography{references.bib}

\addtolength{\textheight}{-12cm}   




\end{document}